\pdfoutput=1

\documentclass[11pt]{article}

\usepackage[]{acl}

\usepackage{times}
\usepackage{latexsym}

\usepackage[T1]{fontenc}

\usepackage[utf8]{inputenc}

\usepackage{microtype}
\usepackage{xcolor}         
\usepackage{threeparttable}  
\usepackage{multirow}
\usepackage{balance}
\usepackage{dsfont}
\usepackage{booktabs}
\usepackage{comment}
\usepackage{enumitem}
\usepackage{graphicx}
\usepackage{enumitem}
\usepackage{amsthm}
\usepackage{amsfonts}
\usepackage{amsmath}
\usepackage{wrapfig}
\usepackage{subfigure}
\usepackage{pifont}
\usepackage{bbm}
\usepackage{comment}
\usepackage{tablefootnote}
\appto\TPTdoTablenotes{\footnotesize}

\theoremstyle{definition}

\newcommand{\re}{\textbf{Re\,}}

\title{KGE-CL: Contrastive Learning of Tensor Decomposition Based Knowledge Graph Embeddings}

\author{Zhiping Luo$^{1,4\ast}$, Wentao  Xu$^{1,4}$\thanks{\ \  The first two authors contributed equally.}, Weiqing Liu$^2$, Jiang Bian$^2$, Jian Yin$^{3,4}$\thanks{\ \ Corresponding author.}, and Tie-Yan Liu$^2$\\
	$^1$ School of Computer Science and Engineering, Sun Yat-sen University, Guangzhou, China\\
	$^2$ Microsoft Research Asia, Beijing, China \\
	$^3$ School of Artificial Intelligence, Sun Yat-sen University, Zhuhai, China\\
    $^4$ Guangdong Key Laboratory of Big Data Analysis and Processing, Guangzhou, China\\
	\texttt{\{luozhp7@mail2,xuwt6@mail2,issjyin@mail\}.sysu.edu.cn} \\
	\texttt{\{weiqing.liu, jiang.bian, tyliu\}@microsoft.com} \\
}

\begin{document}
	
	\maketitle
	
	\begin{abstract}
Learning the embeddings of knowledge graphs (KG) is vital in artificial intelligence, and can benefit various downstream applications, such as recommendation and question answering.
In recent years, many research efforts have been proposed for knowledge graph embedding (KGE).
However, most previous KGE methods ignore the semantic similarity between the related entities and entity-relation couples in different triples since they separately optimize each triple with the scoring function.
To address this problem, we propose a simple yet efficient contrastive learning framework for tensor decomposition based (TDB) KGE, which can shorten the semantic distance of the related entities and entity-relation couples in different triples and thus improve the performance of KGE.
We evaluate our proposed method on three standard KGE datasets: WN18RR, FB15k-237 and YAGO3-10.
Our method can yield some new state-of-the-art results, achieving 51.2\% MRR, 46.8\% Hits@1 on the WN18RR dataset, 37.8\% MRR, 28.6\% Hits@1 on FB15k-237 dataset, and 59.1\% MRR, 51.8\% Hits@1 on the YAGO3-10 dataset.
Source codes and data of this paper can be found at \url{https://github.com/Wentao-Xu/KGE-CL}.
\end{abstract}

	\maketitle

	\section{Introduction}
\label{sec:intro}
The knowledge graph (KG) stores a vast number of human knowledge in the format of triples.
A triple $(h, r, t)$ in a KG contains a head entity $h$, a tail entity $t$, and a relation $r$ between $h$ and $t$.
The knowledge graph embedding (KGE) aims to project the massive interconnected entities and relations in a KG into vectors or matrices, which can preserve the semantic information of the triples.
Learning the embeddings of KG can benefit various downstream artificial intelligence applications, such as question answering~\cite{huang2019knowledge}, machine reading comprehension~\cite{yang-mitchell-2017-leveraging}, image classification~\cite{marino2016more}, and personalized recommendation~\cite{wang2018dkn}. 

\begin{figure*}[t]
	\centering
	\includegraphics[width=1.5\columnwidth]{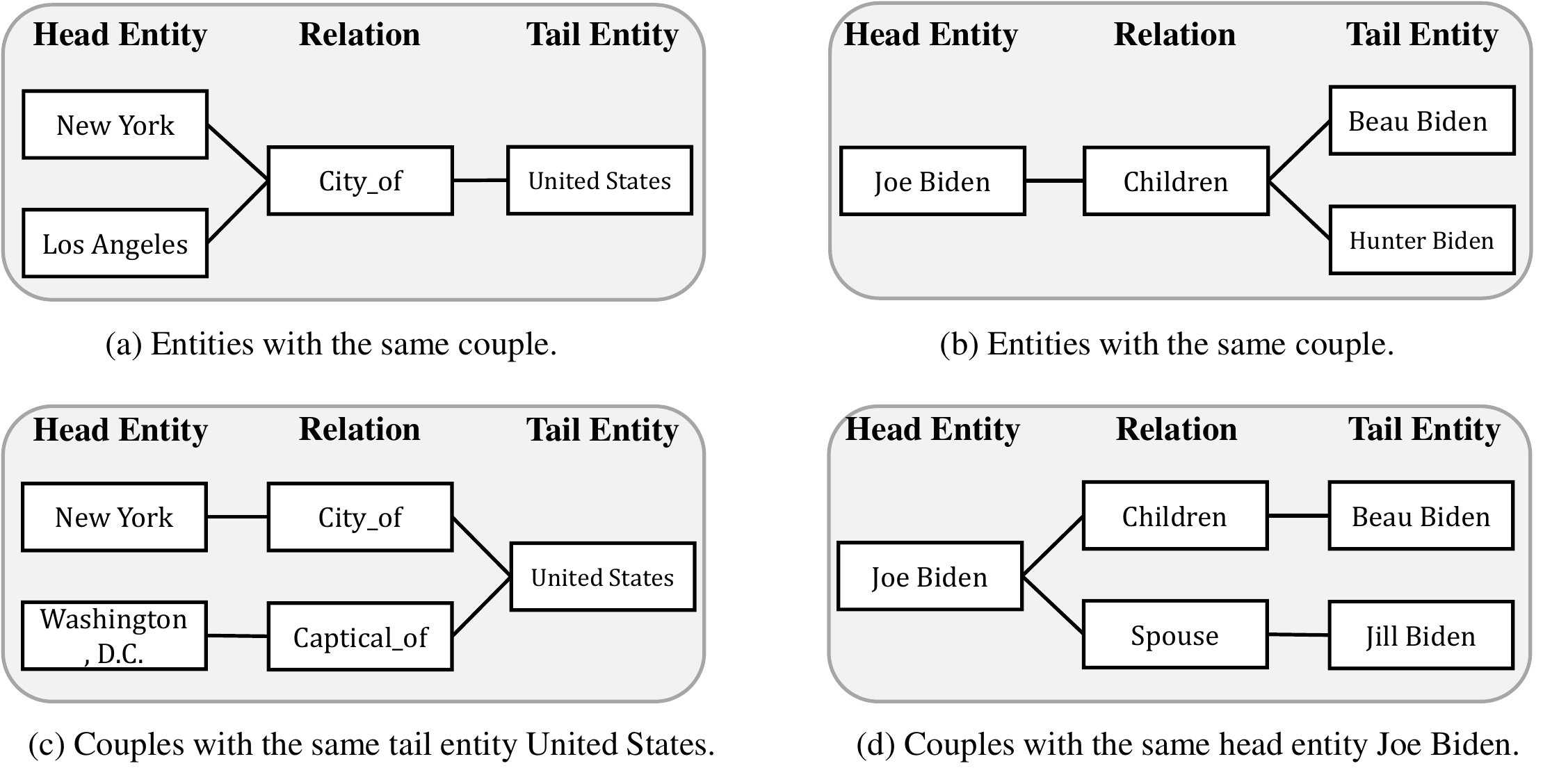}
	\caption{The examples of triples that share the same entities or entity-relation couples.}
	\label{fig:example}
\end{figure*} 

In general, most of the KGE methods would define a scoring function $f(h_i,r_j,t_k)$, and the training target of KGE is maximizing the score of a true triple $(h_i, r_j, t_k)$ and minimizing the score of a false tripe $(h_i, r_j, t_x)$. 
In this way, the trained embeddings of entities and relations in KG can preserve the intrinsic semantics of a true triple. 
We can mainly divide the existing KGE methods into two categories. The first category is the distance based (DB) methods, which use the Minkowski distance as scoring function to measure a triple's plausibility, include the TransE~\cite{transe}, TransH~\cite{transh}, TransR~\cite{transr}, TransD~\cite{transd} and TransG~\cite{transg}. The other category is the tensor decomposition based (TDB) methods, which treat a KG as a third-order binary tensor and use the results of tensor decomposition as the representations of entities and relations. The TDB methods include the CP~\cite{cp}, DistMult~\cite{distmult}, RESCAL~\cite{rescal} and ComplEx~\cite{complex}.

However, existing methods only capture the semantic connections among $h$, $r$ and $t$ in a same triple. 
For example, the TransE optimize the distance between $h+r$ and $t$, and the DistMult optimizie the dot product similarity among $h$, $r$ and $t$.
Therefore, they overlook the connections between the related entities and entity-relation couples in different triples. 
In a KG, some entities (entity-relation couples) that share the same entity-relation couple (entity) are in the same type and have semantic similarity. Capturing the semantic similarity of these entities or couples can improve the expressiveness of embeddings, which is the performance in capturing semantic information in KG~\cite{conve,seek}.
For instance, in Figure~\ref{fig:example} (a), the entities \textit{New York} and \textit{Los Angeles} share the same entity-relation couple (\textit{City\_of}, \textit{United States}). 
Therefore, the entities \textit{New York} and \textit{Los Angeles} should have a similar embedding.
Besides, in Figure~\ref{fig:example} (c) the couples (\textit{New York}, \textit{City\_of}) and (\textit{Washington, D.C.}, \textit{Captical\_of}) share the same tail entity \textit{United States}. Thus the representations of  these two couples should also be similar.

To correlate the related entities and entity-relation couples in different triples, we propose a simple yet efficient contrastive learning framework called KGE-CL for KGE, which is quite general for existing TDB methods.
We first construct the positive instances for those entities that share the same entity-relation couple and those entity-relation couples that share the same entity.
For example, the positive instance of \textit{Beau Biden} in Figure~\ref{fig:example} (b) is the \textit{Hunter Biden}, and the positive instance of (\textit{Children}, \textit{Beau Biden}) in Figure~\ref{fig:example} (d) is (\textit{Spouse}, \textit{Jill Biden}).
Then we calculate the contrastive loss on the original instance and the positive instances.
Due to the design of the contrastive learning framework, we can also increase the distance between unrelated entities and couples.
Finally, since each triple has four positive instances (corresponding to the four examples in Figure~\ref{fig:example}), we design a weighted contrastive loss to control the weights on different positive instances' loss flexibly.

We evaluate our KGE-CL method on the KG link prediction task using the standard WN18RR~\cite{wn18rr}, FB15k-237~\cite{conve} and YAGO3-10~\cite{yago3} datasets.
Our proposed method achieves new state-of-the-art results (SotA), obtaining 51.2\% MRR, 46.8\% Hits@1 on the WN18RR dataset, 37.8\% MRR, 28.6\% Hits@1 on the FB15k-237 dataset, and 59.1\% MRR, 51.8\% Hits@1 on the YAGO3-10 dataset.
Moreover, We apply several experiments to further analyze the inner mechanism of our method.
At last, to clearly explain why our method outperforms existing methods, we conduct the visualization of the KGE of our method and some compared methods using T-SNE~\cite{tsne}.

In summary, this paper's contributions include:
\begin{itemize}[leftmargin=1.5em]
	\item We propose KGE-CL, a simple yet efficient contrastive learning framework for TDB KGE. It can capture the semantic similarity of the related entities and entity-relation couples in different triples, thus improving the performance of KGE. 
	\item Our proposed KGE-CL framework can also push the embeddings of unrelated entities and couples apart in the semantic space.
	\item  The experiment results and analyses confirm the effectiveness of our KGE-CL method.
\end{itemize}

	\section{Related Work}\label{sec:related_work}
In recent years, knowledge graph embedding (KGE) becomes a pretty hot research topic since its vital role in various downstream applications.
We can categorize the existing KGE techniques into two categories: the distance based KGE and tensor decomposition based KGE.

Distance based (DB) methods describe relations as relational maps between head and tail entities. 
The TransE is a representative distance based method, which uses the relations as translations and its scoring function is: $f(h_i, r_j, t_k) = -||\textbf{h}_i+\textbf{r}_j - \textbf{t}_k||_2^2$.
To improve the performance of TransE, many its variants that follow the same direction were proposed, such as the TransH~\cite{transh}, TransR~\cite{transr}, TransD~\cite{transd}, TranSparse~\cite{transparse}, TransG~\cite{transg} and RotatE \cite{rotate}.
However, the TransE and its extensions can not capture the semantic similarity between the related entities and entity-relation couples in different triples. For example, given two triples $(h_1, r_1, t_1)$ and $(h_1, r_1, t_2)$, TransE can only close the distance between $\textbf{h}_1 + \textbf{r}_1$ and $\textbf{t}_1$ (or $\textbf{t}_2$) in the same triple, it does not close the distance between $\textbf{t}_1$ and $\textbf{t}_2$, so the representations $\textbf{t}_1$ and $\textbf{t}_2$ can in different directions. 

Tensor decomposition based (TDB) methods formulate the KGE task as a third-order binary tensor decomposition problem. 
RESCAL \cite{rescal} factorizes the $j$-th frontal slice of $\mathcal{X}$ as $\mathcal{X}_j\approx \textbf{A}\textbf{R}_j\textbf{A}^\top$, in which embeddings of head and tail entities are from the same space. As the relation embeddings in RESCAL are matrices containing lots of parameters, RESCAL is easier to be overfitting and more difficult to train. DistMult \cite{distmult} simplifies the matrix $\textbf{R}_j$ in RESCAL to a diagonal matrix, while the RESCAL can only preserve the symmetry of relations, limiting its expressiveness. 
To model asymmetric relations, ComplEx \cite{complex} extends DistMult to complex embeddings and preserving the relations' symmetry in the real part and the asymmetry in the imaginary part.
Moreover, the QuatE~\cite{quate} further extends the ComplEx to hypercomplex space to model more complicated relation properties, such as the inversion.
All of the DistMult, ComplEx, and QuatE are the variants of CP decomposition \cite{cp}, which are in real, complex, and hypercomplex vector spaces, respectively.
On the other hand, the TDB methods usually suffer from an overfitting problem; thus, some work is trying to improve the TDB methods from the aspect of regularizer, such as the N3~\cite{n3} and DURA~\cite{dura} regularizers. These regularizers bring more significant improvements than the original squared Frobenius norm ($L_2$ norm) regularizer~\cite{rescal,distmult,complex}.
Nevertheless, the TDB methods can only capture the similarity among $h$, $r$ and $t$ in a same triple. For example, the ComplEx-DURA~\cite{dura} mainly capture the semantics similarity between the couple $(h_i, r_j)$'s embedding $\textbf{h}_i\textbf{R}_j$ and tail entity $t_k$'s embedding $\textbf{t}_k$ in a sample triple. 

Since both of the DB and TDB methods can not correctly capture the semantic similarity between related entities and couples in different triples, we propose our KGE-CL method  to address the limitations of existing work.

	\section{Preliminaries}
\subsection{Knowledge Graph Embedding}\label{sec:kge}

\noindent\textbf{Knowledge Graph Embedding (KGE)} 
The knowledge graph embedding (KGE) is to learn the representations (may be real or complex vectors, matrices, and tensors) of the entities and relations. Its target is that the learned entities' and relations' embeddings can preserve the semantic information of the triples in knowledge graphs.
Generally, the KGE methods define a scoring function$f(h_i, r_j, t_k)$ to score the corresponding triple $(h_i, r_j, t_k)$, and the score measure the plausibility of triples.


\noindent\textbf{Tensor Decomposition Based (TDB) KGE} 
TDB methods like RESCAL~\cite{rescal} and ComplEx~\cite{complex}, regard a KG as a third-order binary tensor $\mathcal{X}\in\{0,1\}^{|\mathcal{E}|\times|\mathcal{R}|\times|\mathcal{E}|}$. The $(i,j,k)$ entry $\mathcal{X}_{ijk}=1$ if $(h_i, r_j, t_k)$ is a true triple otherwise $\mathcal{X}_{ijk}=0$. 
The $\mathcal{X}_j$ denotes the $j$-th frontal slice of $\mathcal{X}$, that is, the corresponding matrix of the $j$-th relation.
Generally, a TDB KGE model factorizes $\mathcal{X}_j$ as $\mathcal{X}_j\approx\re(\textbf{H} \textbf{R}_j\overline{\textbf{T}}^\top)$, where the $i$-th ($k$-th) row of $\textbf{H}$ ($\textbf{T}$) is $\textbf{h}_i$ ($\textbf{t}_k$), $\textbf{R}_j$ is a matrix that represents relation $r_j$, $\re(\cdot)$ and $\overline{\cdot}$ are the real part and the conjugate of a complex matrix, respectively. 
Then the scoring functions of TDB KGE methods is: $f(h_i,r_j,t_k)=\re(\textbf{h}_i\textbf{R}_j\overline{\textbf{t}}_k^\top)$. Note that the real part and the conjugate of a real matrix are itself. 
The goal of TDB models is to seek matrices $\textbf{H}, \textbf{R}_1,\dots,\textbf{R}_{|\mathcal{R}|},\textbf{T}$, such that $\re(\textbf{H} \textbf{R}_j\overline{\textbf{T}}^\top)$ can approximate $\mathcal{X}_j$.
In this paper, we aim to improve the performance of existing TDB models, such as the RESCAL and ComplEx models.

\subsection{Contrastive Learning}\label{sec:b_cl}
Contrastive learning is an efficient representation learning method that contrasts positive pairs against negative pairs~\cite{hadsell2006dimensionality,he2020momentum, simclr,khosla2020supervised}.
The key idea of contrastive learning is pulling the semantically close pairs together and push apart the negative pairs.
The unsupervised contrastive learning framework~\cite{simclr} would utilize the data augmentation to construct positive pairs to calculate the contrastive loss.
The supervised contrastive learning framework~\cite{khosla2020supervised} calculates the contrastive loss of all positive instances within the same mini-batch.
Motivated by these exiting frameworks, we adopt the following function to calculate the contrastive loss between an instance $z_i$ and its all positive instances $z_i^+$:
\begin{equation}
	\label{eq:cl}
	\mathrm{CL}\left(
	\textbf{z}_i,  \textbf{z}_i^+\right) =\frac {-\sum_{z_i^+\in P(i)} \log \frac{ e^{\mathrm{sim}\left( \textbf{z}_i,  \textbf{z}_i^+\right)/\tau}}{\sum_{z_j \in N(i)}e^{\mathrm{sim}\left(\textbf{z}_i,  \textbf{z}_j \right)/\tau}}}{\mid P(i) \mid},
\end{equation}
where $\textbf{z}_i$ and $\textbf{z}_i^+$ are the representations of $z_i$ and $z_i^+$, respectively.
$P(i)$ is the set of all positive instances in the mini-batch, and $N(i)$ is the set of all negative instances in the batch. In our work, we define the negative instances as the instances that do not belong to the positive instances.
$\mathrm{sim}(\textbf{z}_i,  \textbf{z}_i^+) = \textbf{z}_i \cdot \textbf{z}_i^+$ is the dot product similarity.

	\section{Our Method: KGE-CL}
\label{sec:method}
In this section, we describe our KGE-CL method that utilize the contrastive learning to capture the semantic similarity of related entities and couples in different triples. 
Our method is very general and can be easily apply to arbitrary TDB methods.
We can further name our KGE-CL method as RESCAL-CL or ComplEx-CL when we use the scoring function of RESCAL or ComplEx models.
In this section, we firstly present the contrastive loss we designed for the KGE, then we introduce the training objective of our method.

\subsection{Contrastive Loss of KGE}
\label{sec:cl}
In this subsection, we elaborate on the contrastive loss that we designed for KGE.
\paragraph{Positive Instances}
The generation of positive instances $z_i^+$ for the instance $z_i$ is vital in contrastive learning.
Existing work in visual representation learning~\cite{wu2018unsupervised,simclr,chen2020exploring} used some data augmentation methods, such as cropping, color distortion, and rotation, to take two random transformations of the same images as $z_i$ and $z_i^+$. Meanwhile, in NLP, some work~\cite{wu2020clear,meng2021coco} utilized other augmentation techniques like word deletion, reordering, and substitution.
However, these data augmentation methods are not proper to the KGE.
To capture the interactions between triples in a KG, we design a new approach to construct the positive instances for KGE.
For a triple $(h_i, r_j, t_k)$, the corresponding scoring function of TDB methods is:
\begin{equation}
	\label{eq:score}
	\begin{split}
		f(h_i, r_j, t_k) = \re(\textbf{h}_i \textbf{R}_j\overline{\textbf{t}}_k^\top) &=\re(\langle \textbf{h}_i \textbf{R}_j, \overline{\textbf{t}}_k\rangle) \\ 
		&=\re(\langle \textbf{h}_i, \overline{\textbf{t}}_k\textbf{R}_j^\top \rangle).
	\end{split}
\end{equation}
We define $\langle \cdot, \cdot \rangle$ as the inner product of two real or complex vectors: $\langle \textbf{u}, \textbf{v} \rangle = \textbf{u}\textbf{v}^{\top}$.
The $\textbf{h}_i \textbf{R}_j$ and $\overline{\textbf{t}}_k\textbf{R}_j^\top$ are the representations of the entity-relation couples $(h_i, r_j)$ and $(r_j, t_k)$, respectively. 
The Equation~\ref{eq:score} means that we can firstly compute either the $\textbf{h}_i \textbf{R}_j$ or $\overline{\textbf{t}}_k\textbf{R}_j^\top$ in the scoring function. 
For a head entity $h_i$, we define its positive instances $h_i^+$ as those head entities that share the same relation and tail entity with $h_i$. 
Similarly, we define the tail entity $t_k$'s positive instances $t_k^+$ with those tail entities that share the same head entity and relation with $t_k$.
For the entity-relation couples $(h_i, r_j)$ or $(r_j, t_k)$, the corresponding positive instances $(h_i, r_j)^+$ or $(r_j, t_k)^+$ is those entity-relation couples that share the same tail entity with $(h_i, r_j)$ or head entity with $(r_j, t_k)$.
The $(\textbf{h}_i \textbf{R}_j)^+$ and $(\overline{\textbf{t}}_k\textbf{R}_j^\top)^+$ are the representations of positive instances $(h_i, r_j)^+$ and $(r_j, t_k)^+$, respectively.
Therefore, given a true triple $(h_i, r_j, t_k)$, our method would construct four kinds of positive instances: $h_i^+$, $t_k^+$, $(h_i, r_j)^+$ and $(r_j, t_k)^+$, which are corresponding to the four examples in Figure~\ref{fig:example}.
As mentioned in Section~\ref{sec:b_cl}, for an instance $h_i$, we will use all of its positive instances $h_i^+$ in the same mini-batch. Besdies, there may be no positive instances in a mini-batch for some tripels. Therefore, we will add a postive instance for each triple in the mini-batch, where the added instance is randomly sampled from training set.  We will study the effect of postive instances by experiment in Section~\ref{sec:model_analysis}.

\paragraph{Contrastive Loss}
Given a true triple $(h_i, r_j, t_k)$, we use the Equation~\ref{eq:cl} to compute the contrastive loss of four types of positive instances, and the overall contrastive loss for a triple $(h_i, r_j, t_k)$ is:
\begin{equation}
	\label{eq:cl_loss}
	\begin{split}
		&\mathcal{L}_c(h_i, r_j, t_k)=\mathrm{CL}(\textbf{h}_i,  \textbf{h}_i^+) + \mathrm{CL}(\textbf{t}_k,  \textbf{t}_k^+)  \\
		&+\mathrm{CL}(\textbf{h}_i \textbf{R}_j,  \left(\textbf{h}_i \textbf{R}_j\right)^+) +\mathrm{CL}(\overline{\textbf{t}}_k\textbf{R}_j^\top,  (\overline{\textbf{t}}_k\textbf{R}_j^\top)^+).
	\end{split}
\end{equation}

\paragraph{Theoretical Analysis}
To study how the Equation~\ref{eq:cl_loss} takes effect, we apply a theoretical analysis from the aspect of gradient.
Taking the contrastive loss term $\mathrm{CL}(\textbf{h}_i,  \textbf{h}_i^+)$ as an example, the gradient of $\mathrm{CL}(\textbf{h}_i,  \textbf{h}_i^+)$ to embedding $\textbf{h}_i$ is:
\begin{equation}
	\begin{aligned}
		\label{eq:loss_gradient}
		&\dfrac{\partial \mathrm{CL}(\textbf{h}_i,  \textbf{h}_i^+)}{\partial{\textbf{h}_i}}= \\
		&\dfrac{\sum_{h_i^+\in P(i)}\partial \left({\textbf{h}_i\cdot\textbf{h}_i^+/\tau}-\log\sum_{h_j\in N(i)} e^{\left(\textbf{h}_i\cdot\textbf{h}_j/\tau\right)}\right)}{-|P(i)| \partial\textbf{h}_i}\\
		&=- \dfrac{ \sum_{h_i^+\in P(i)}\textbf{h}_i^+}{\tau|P(i)|} + \dfrac{ \sum_{h_j \in N(i)} \left( e^{\left(\textbf{h}_i\cdot\textbf{h}_j /\tau\right)}\textbf{h}_j\right)}{\tau \sum_{h_j\in N(i)}e^{\left(\textbf{h}_i\cdot\textbf{h}_j/\tau\right)}}.
	\end{aligned}
\end{equation}
Then when we update the embedding $\textbf{h}_i$ with the gradient $\dfrac{\partial \mathrm{CL}(\textbf{h}_i,  \textbf{h}_i^+)}{\partial{\textbf{h}_i}}$:
\begin{equation}
\label{eq:update}
\textbf{h}_i^{t+1} = \textbf{h}_i^t - \eta \dfrac{\partial \mathrm{CL}(\textbf{h}_i,  \textbf{h}_i^+)}{\partial{\textbf{h}_i}}, \\
\end{equation}
where $\eta$ of learning rate for updating gradient. The Equation~\ref{eq:loss_gradient} and Equation~\ref{eq:update} show that the embedding $\textbf{h}_i^t$ would update to the direction of $\dfrac{\eta \sum_{h_i^+\in P(i)}\textbf{h}_i^+}{\tau|P(i)|}$, which is the mean value of positive instances' embeddings $\textbf{h}_i^+$. Meanwhile, the $\textbf{h}_i^t$ would also update away from the weighted value $\dfrac{\eta \sum_{h_j \in N(i)} \left( e^{\left(\textbf{h}_i\cdot\textbf{h}_j /\tau\right)}\textbf{h}_j\right)}{\tau \sum_{h_j\in N(i)}e^{\left(\textbf{h}_i\cdot\textbf{h}_j/\tau\right)}}$ of negative instances $\textbf{h}_j$.
So our proposed contrastive loss $\mathcal{L}_c$ can not only pull the related entities and entity-relation couples together in the semantic space but also push the unrelated entities and couples apart.

\begin{table}[t]
	\centering
	\begin{threeparttable}
		\resizebox{\columnwidth}{!}{
			\begin{tabular}{l| c c c  c |c c c c }
				\toprule
				\multirow{2}{*}{\textbf{Contrastive Loss}} &\multicolumn{4}{c}{\textbf{WN18RR}}&  \multicolumn{4}{c}{\textbf{FB15k-237}}\\
				\cmidrule(lr){2-5}
				\cmidrule(lr){6-9}
				& MRR & H@1 & H@3 & H@10& MRR & H@1 & H@3 & H@10 \\
				\midrule
				\midrule
				$\mathrm{CL}(\textbf{h}_i,  \textbf{h}_i^+)$      &.509 &.465 &\textbf{.527}&.588 &.376& .284& .412& .557\\
				$\mathrm{CL}(\textbf{t}_k,  \textbf{t}_k^+)$  &.509 &.468 &.425&.587&.376&.285&.411&.558  \\
				$\mathrm{CL}(\textbf{h}_i \textbf{R}_j,  (\textbf{h}_i \textbf{R}_j)^+)$  &\textbf{.512} &\textbf{.469} &\textbf{.527} &\textbf{.595}&.374&.282&.410&.557  \\
				$\mathrm{CL}(\overline{\textbf{t}}_k\textbf{R}_j^\top,  (\overline{\textbf{t}}_k\textbf{R}_j^\top)^+)$ & .504  & .462 &.516& .580 &\textbf{.378}&\textbf{.286}&\textbf{.414}&\textbf{.559} \\
				\bottomrule
			\end{tabular}
		}
	\end{threeparttable}
	\caption{Link prediction results of RESCAL-CL that merely uses one specific contrastive loss term.}
	\label{table:observation}
\end{table}

\paragraph{Weighted Contrastive Loss}
There are four contrastive loss terms in Equation~\ref{eq:cl_loss}. We found that different contrastive loss terms have different effects on different knowledge graphs during our research process. 
This phenomenon happens may because different knowledge graphs have diverse graph properties, such as the ratio of the number of entities to the number of relations, the number of triples compared with the number of entities.
Table~\ref{table:observation} shows the results of RESCAL-CL that merely uses one specific contrastive loss term. 
In WN18RR dataset, using the term $\mathrm{CL}(\textbf{h}_i \textbf{R}_j,  (\textbf{h}_i \textbf{R}_j)^+)$ achieves the highest results, while in FB15k-237 dataset, $\mathrm{CL}(\overline{\textbf{t}}_k\textbf{R}_j^\top,  (\overline{\textbf{t}}_k\textbf{R}_j^\top)^+)$ is the best.
Hence, we introduce a weighted contrastive loss, assigning a weight $\alpha_*$ for each contrastive loss term, and $\alpha_*$ is a hyper-parameter that can be flexibly tuned for a specific KG. 
The contrastive loss $\mathcal{L}_c^w(h_i, r_j, t_k)$ after adding weights $\alpha_*$ is:
\begin{equation}
	\label{eq:cl_wloss}
	\begin{split}
		\mathcal{L}_c^w(h_i, r_j, t_k)=&\alpha_h \mathrm{CL}(\textbf{h}_i,  \textbf{h}_i^+) + \alpha_t\mathrm{CL}(\textbf{t}_k,  \textbf{t}_k^+)  \\
		+&\alpha_{hr}\mathrm{CL}(\textbf{h}_i \textbf{R}_j,  (\textbf{h}_i \textbf{R}_j)^+) \\ 
		+&\alpha_{tr}\mathrm{CL}(\overline{\textbf{t}}_k\textbf{R}_j^\top,  (\overline{\textbf{t}}_k\textbf{R}_j^\top)^+).
	\end{split}
\end{equation}

\subsection{Training Objective}
Given a training triple $(h_i, r_j, t_k)$ in a KG, the instantaneous loss of our framework on this triple is:
\begin{equation}
	\label{eq:objective}
	\begin{split}
		\mathcal{L}(h_i, r_j, t_k) =  \mathcal{L}_s + \mathcal{L}_r +\mathcal{L}_c^w,
	\end{split}
\end{equation}
where $\mathcal{L}_s$ is the loss that measures the discrepancy between scoring function's output $f(h_i,r_j,t_k)$ and the label $\mathcal{X}_{ijk}$.
$\mathcal{L}_r$ is the regularizer, and $\mathcal{L}_c^w$ is the weighted contrastive loss we introduced in Section~\ref{sec:cl}.
It should be noted that the additionally added postive instances in the mini-batch is only used to calculate the contrastive loss $\mathcal{L}_c^w$ and would not used to calculate the $\mathcal{L}_s$ and $\mathcal{L}_r$ losses.

\paragraph{$\mathcal{L}_s$ Loss}
Many previous efforts used the ranking losses~\cite{transe}, binary logistic regression~\cite{complex} or sampled multiclass log-loss~\cite{kadlec2017knowledge} to calculate the distance between the scoring function's output and the triple's label.
Since~\cite{n3} had verified the competitiveness of the full multiclass log-loss, we utilize it as the $\mathcal{L}_s$ loss in Equation~\ref{eq:objective}.
\paragraph{Regularizer}
Most of the previous work use the squared Frobenius norm ($L_2$ norm) regularizer~\cite{rescal,distmult,complex} in their object functions. More recently, some work proposed more efficient regularizers to prevent the overfitting of KGE, such as the N3~\cite{n3} and DURA~\cite{dura} regularizers. 
Since the~\cite{dura} had shown that DURA regularizer outperforms L2 and N3 regularizers, we use the DURA as the regularizer $\mathcal{L}_r$ in our work.
	\section{Experiment}
\label{sec:experiment}

In this section, we present thorough empirical studies to evaluate and analyze our proposed framework. We first introduce the experimental setting. Then we evaluate our framework's performance on the task of link prediction.
Besides, we further analyze the details of our promotion by comparing our method with a baseline on the triples with different relations, and we also study the effect of positive instances to our framework.
Finally, we visualize the embeddings of our method and some baselines to explain why our method outperforms baselines.
\begin{table}[t]
	\centering
	\begin{threeparttable}
		\resizebox{\columnwidth}{!}{
			\begin{tabular}{l r r r}
				\toprule
				& WN18RR & FB15k-237 & YAGO3-10 \\
				\midrule
				\#Entity & 40,943 & 14,541 & 123,182\\
				\#Relation & 11 & 237 & 37 \\
				\#Train & 86,835 & 272,115 & 1,079,040 \\
				\#Valid &3,034 &17,535 &5,000\\
				\#Test &3,134 &20,466 &5,000\\
				\bottomrule
			\end{tabular}
		}
		\caption{Statistics of the datasets.}
		\label{table:datasets}
	\end{threeparttable}
\end{table}
\subsection{Exeprimental Setting}
\paragraph{Dataset}
We use three standard KG datasets---WN18RR \cite{wn18rr}, FB15k-237 \cite{conve}, and YAGO3-10 \cite{yago3} to evaluate the performance of KGE. We divide the datasets into training, validating, and testing sets using the same way of previous work. Table \ref{table:datasets} shows the statistics of these datasets.
WN18RR, FB15k-237, and YAGO3-10 are extracted from WN18 \cite{transe}, FB15k \cite{transe}, and YAGO3 \cite{yago3}, respectively. Some previous work~\cite{wn18rr,conve} indicated the test set leakage problem in WN18 and FB15k, where some test triplets may appear in the training dataset in the form of reciprocal relations. Therefore, they suggested using the WN18RR and FB15k-237 datasets to avoid the test set leakage problem.

\paragraph{Compared Methods} 
We compare our KGE-CL method with existing state-of-the-art KGE methods, including CP~\cite{cp}, RESCAL~\cite{rescal}, ComplEx~\cite{complex}, ConvE~\cite{conve}, RoratE~\cite{rotate}, MuRP~\cite{murp}, HAKE~\cite{hake}, ComplEx-N3~\cite{n3}, ROTH~\cite{roth}, REFE~\cite{roth}, CP-DURA~\cite{dura}, RESCAL-DURA~\cite{dura} and ComplEx-DURA~\cite{dura}. 

\begin{table}[t]
	\centering
	\begin{threeparttable}
		
		\resizebox{\columnwidth}{!}{
			\begin{tabular}{l|c| c  c c c c c c}
				\toprule
				\textbf{Datasets} &\textbf{Methods} & $ d $  & $ m $ & $\tau $ & $\alpha_h$ & $\alpha_t$ & $\alpha_{hr}$ & $\alpha_{tr}$\\
				\midrule
				\midrule
				\multirow{2}{*}{\textbf{WN18RR}}&RESCAL-CL &512    &512 & 0.9 & 0 &  0& 2.0 &0 \\
				&ComplEx-CL&2000   & 2048 & 0.5 & 0 & 0 & 0 & 2.0\\
				\midrule
				\multirow{2}{*}{\textbf{FB15k-237}}&RESCAL-CL &512   &512 & 0.9 & 0 & 0 &0 &2.0\\
				&ComplEx-CL& 2000  &2048 &0.5 & 2.0 & 0 &0 &0\\
				\midrule
				\multirow{2}{*}{\textbf{YAGO3-10}}&RESCAL-CL &512    &512 & 0.9 &  0& 0& 0&1.0\\
				&ComplEx-CL& 2000 &2048 &0.5 & 0 & 1.0 & 0 & 0\\
				\bottomrule
			\end{tabular}
		}
	\end{threeparttable}
	\caption{The selection of the hyper-parameters of RESCAL-CL and ComplEx-CL on different datasets.}
	\label{table:alpha}
\end{table}
\begin{table*}[t]
	\centering
	\begin{threeparttable}
		\resizebox{0.96\textwidth}{!}{
			\begin{tabular}{l | c c c  c | c c  c c | c c c c }
				\toprule
				\multirow{2}{*}{\textbf{Methods}} &\multicolumn{4}{c}{\textbf{WN18RR}}&  \multicolumn{4}{c}{\textbf{FB15k-237}} & \multicolumn{4}{c}{\textbf{YAGO3-10}}\\
				\cmidrule(lr){2-5}
				\cmidrule(lr){6-9}
				\cmidrule(lr){10-13}
				& MRR & H@1 & H@3 & H@10 & MRR & H@1 & H@3  & H@10 & MRR & H@1 & H@3  & H@10 \\
				\midrule
				\midrule
				CP      &.438 &.414 &.445&.485 &.333 &.247 &.363&.508 &.567 &.494 &.611&.698\\
				RESCAL  &.455 &.419 &.461&.493 &.353 &.264 &.385&.528 &.566 &.490 &.612&.701\\
				ComplEx & .460  & .428 &.473& .522  & .346 & .256 &.386& .525 & .573 & .500 &.617& .703\\
				ConvE  &.43 &.40 &.44 &.52 &.325 &.237& .356 &.501 &.44 &.35 &.49& .62 \\
				RotatE & .476 & .428 &.492 & .571 &  .338 &  .241 &.375& .533 & .495  &  .402 &.550& .670\\
				MuRP & .481 & .440 & .495 &.566 & .335 & .243 &.367&.518 &- &- &&-\\
				HAKE & .497 &.452 & .516 & .582 &.346  &.250 &.381&.542 & .546  & .462 &.596&.694\\
				ComplEx-N3 &.491 &.448 &.505&.580 &.366 &.271 &.403&.558 &.577 &.502 &.619&.711\\
				ROTH &.496 &.449 &.514 &.586 &.344 &.246 &.380 &.535 &.570 &.495 &.612 &.706\\
				REFE &.473 &.430 &.485 &.561 &.351 &.256 &.390 &.541 &.577 &.503 &.621 &.712\\
				CP-DURA  &.478 &.441 &.497&.552 &.367  &.272 &.402&.555 &.582 &.511 &.623&.708\\
				RESCAL-DURA    &.498 &.455 &.514&.577 &.368 &.276 &.402 &.550 &.579 &.505 &.619&.712\\
				ComplEx-DURA &.491 &.449 &.504&.571 &.371 &.276 &.408&.560 &.584 &.511 &.628&.713\\
				\midrule
				\textbf{RESCAL-CL}&\textbf{.512} &\textbf{.468} &\textbf{.531}&\textbf{.597} & \textbf{.378}&\textbf{.286}&\textbf{.414}&.559&.581 &.507 &.625&.713\\
				\textbf{ComplEx-CL} &.505 &.458 &.522&.595&.377&.285 &\textbf{.414} &\textbf{.564}&\textbf{.591} &\textbf{.518} &\textbf{.634} &\textbf{.722}\\
				\bottomrule
			\end{tabular}
		}
	\end{threeparttable}
	\caption{Link prediction results on WN18RR, FB15k-237 and YAGO3-10 datasets. We take the results of CP, RESCAL, ComplEx,  CP-DURA, RESCAL-DURA and ComplEx-DURA from the paper~\cite{dura}, and the results of other baselines are from their original papers.}
	\label{table:main_results}
\end{table*}
\paragraph {Implementation Details}
We implement our method base on the PyTorch library~\cite{paszke2019pytorch}, and run on all experiments with a single NVIDIA Tesla V100 GPU. We leverage Adagrad algorithm~\cite{adagrad} to optimize the objective function in Equation~\ref{eq:objective}.
We tune our model using the grid search to select the optimal hyper-parameters based on the performance on the validation dataset. We search the embedding size $d$ in \{256, 512, 1024\} for RESCAL-CL and \{200, 500, 1000, 2000\} for ComplEx-CL. 
We search the temperature $\tau$ in Equation~\ref{eq:cl} in \{0.3, 0.5, 0.7, 0.9, 1.0\}. 
We search the weights  $\alpha_h$, $\alpha_t$, $\alpha_{hr}$ and $\alpha_{tr}$ in Equation~\ref{eq:cl_wloss} in \{0, 0.2, 0.4, 0.6, 0.8, 1.0, 1.5, 2.0, 2.5\}. 
The best choices of hyper-parameters, the number of parameters, and the training time of RESCAL-CL and ComplEx-CL on each dataset are listed in Table~\ref{table:alpha}. 
For a fair comparison, the RESCAL-CL and ComplEx-CL have the same embedding size as RESCAL-DURA and ComplEx-DURA, respectively.
Besides, the batch size is 512 for RESCAL-CL and 200 for ComplEx-CL, and the learning rate $\eta$ as 0.1 for all methods. 
On WN18RR, we set the number of training epochs as 50 for the ComplEx-CL and 200 for the RESCAL-CL.
On FB15k-237 and YAGO3-10, the number of training epochs is 200 for all methods.

\subsection{Main Results}
We evaluate the performance of our framework on the link prediction task, which is a frequently-used task to evaluate the KGE.
Specifically, we replace the head or tail entity of a true triple in the test set with other entities in the dataset and name these derived triples as \textit{corrupted triples}. The link prediction task aims to score the original true triples higher than the corrupted ones. We rank the triples by the results of the scoring function.

The evaluation metrics we used in the link prediction are the MRR and Hits@N: 1) MRR: the mean reciprocal rank of original triples; 2) Hits@N: the percentage rate of original triples ranked at the top $N$ in prediction.
For both metrics, we remove some of the corrupted triples that exist in datasets from the ranking results, which is also called \emph{filtered} setting in~\cite{transe}. For the metrics of Hits@N, we use Hits@1, Hits@3, and Hits@10.

Table~\ref{table:main_results} shows the results of link prediction on WN18RR, FB15K-237, and YAGO3-10 datasets. 
Our proposed method achieves the highest results on all datasets compared with the baselines.
Specifically, the RESCAL-CL achieves evidently better results on the WN18RR dataset. The ComplEx-CL outperforms the compared methods in the YAGO3-10 dataset.
On FB15k-237, the RESCAL-CL and ComplEx-CL are both better than the RESCAL-DURA and ComplEx-DURA, respectively. 
The results of RESCAL-CL and ComplEx-CL verify that correlating the entities and entity-relation couples in different triples can boost the performance of KGE.

\begin{table}[t]
	\centering
	\begin{threeparttable}
		\resizebox{\columnwidth}{!}{
			\begin{tabular}{l |r|r|ccc|ccc}
				\toprule
				\multirow{2}{*}{\textbf{Relations}}  & \multirow{2}{*}{\textbf{\#Train}}& \multirow{2}{*}{\textbf{\#Test}}&\multicolumn{3}{c}{\textbf{RESCAL-DURA}}&  \multicolumn{3}{c}{\textbf{RESCAL-CL}}\\
				\cmidrule(lr){4-6}
				\cmidrule(lr){6-9}
				&&& MRR & H@1 & H@10 & MRR & H@1& H@10\\
				\midrule
				\midrule
				\_similar\_to  & 80  & 3  & 0.446    & 0.333       & 0.667   & \textbf{0.756} & \textbf{0.667}  & \textbf{1.000} \\
				\_verb\_group   & 1138  & 39  & 0.930    & 0.885        & 0.974   & 0.934          & \textbf{0.897}           & 0.974          \\
				*domain\_usage & 629& 24    & 0.400    & 0.354       & 0.542   & \textbf{0.447} & \textbf{0.396} & 0.542          \\
				*domain\_region & 923 & 26  & 0.329    & 0.269       & 0.442   & \textbf{0.360} & \textbf{0.289}  & \textbf{0.500} \\
				\_member\_meronym & 7402   & 253 & 0.251    & 0.164      & 0.415   & \textbf{0.318} & \textbf{0.221} & \textbf{0.506} \\
				\_has\_part    & 4816 & 172 & 0.223    & 0.151        & 0.375   & \textbf{0.245} & \textbf{0.174}  & \textbf{0.384} \\
				\_hypernym  & 34796  & 1251  & 0.193    & 0.140      & 0.288   & \textbf{0.204} & \textbf{0.152} & \textbf{0.296} \\
				\_instance\_hypernym  & 2921 & 122  & 0.431    & 0.348     & 0.603   & \textbf{0.461} & \textbf{0.369}  & \textbf{0.631} \\
				\_synset\_domain*  & 3116  & 114 & 0.405    & 0.347      & 0.522   & \textbf{0.444} & \textbf{0.395}  & \textbf{0.544} \\
				*related\_form & 29715 & 1074 & 0.957    & 0.951      & 0.967   & 0.959          & 0.954                  & 0.969          \\
				\_also\_see  & 1299 & 56 & 0.606    & 0.554      & 0.679   & \textbf{0.621} & \textbf{0.571}          & \textbf{0.696} \\
				\bottomrule
			\end{tabular}
		}
	\end{threeparttable}
	\caption{MRR, Hit@1 and Hit@10 results of RESCAL-DURA and RESCAL-CL methods on the triples with different relations in WN18RR dataset. We use * to represent the abbreviation of some words in the relation names. \#Train and \#Test are the number of triples with the corresponding relations in the training and test set.}
	\label{table:improve}
\end{table}

\subsection{Model Analysis}
\label{sec:model_analysis}
\paragraph{Analyzing the Improvements}
To further explore why our method outperforms existing state-of-the-art techniques, we compare our RESCAL-CL method with the RESCAL-DURA on the triples with different relations in WN18RR.
Table~\ref{table:improve} shows the results of the comparison, and we found that our RESCAL-CL is significantly better than the RESCAL-DURA in 9 out of the 11 relations, verifying that the promotion of our framework is extensive and not just on some specific relations.

\begin{table}[t]
	\begin{threeparttable}
		\resizebox{\columnwidth}{!}{
			\begin{tabular}{l c|cc|cc}
				\toprule
				\multirow{2}{*}{Methods}&\multirow{2}{*}{Variants}  & \multicolumn{2}{c}{WN18RR} & \multicolumn{2}{c}{FB15k-237} \\
				\cmidrule(lr){3-6}
				\multicolumn{1}{c}{}      &      \multicolumn{1}{c}{}              & MRR   & Hit@10   & MRR      & Hit@10    \\
				\midrule
				\midrule
			   RESCAL &w/o Pos  &0.501  & 0.581 & 0.368 & 0.551\\
				-CL&w/ Pos  &\textbf{0.512}  & \textbf{0.597}     &\textbf{0.378}  & \textbf{0.559}   \\
				\midrule
				\midrule
				ComplEx &w/o Pos   & 0.493 & 0.579 &0.370  &0.560 \\
			   -CL&	w/ Pos       &\textbf{0.505}  & \textbf{0.595}     &\textbf{0.377}  & \textbf{0.564}   \\
				\bottomrule    
			\end{tabular}
		}
	\end{threeparttable}
	\caption{Effect of Positive Instances.}
	\label{table:posivitve_instance}
\end{table}

\begin{figure*}[t]
	\centering
	\subfigure[RESCAL]{\includegraphics[width=0.31\textwidth]{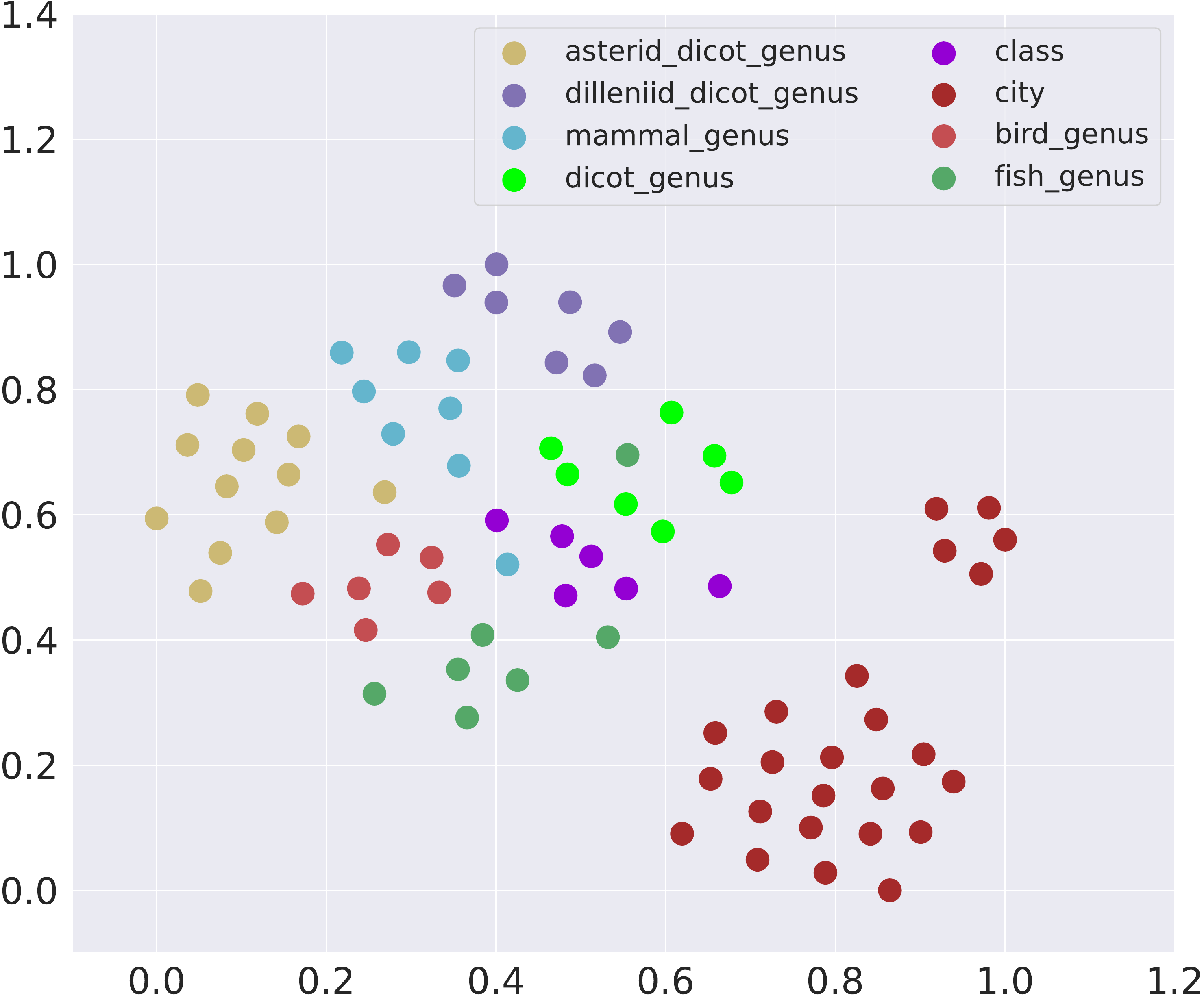}}
	\hspace{.1in} 
	\subfigure[RESCAL-DURA]{\includegraphics[width=0.31\textwidth]{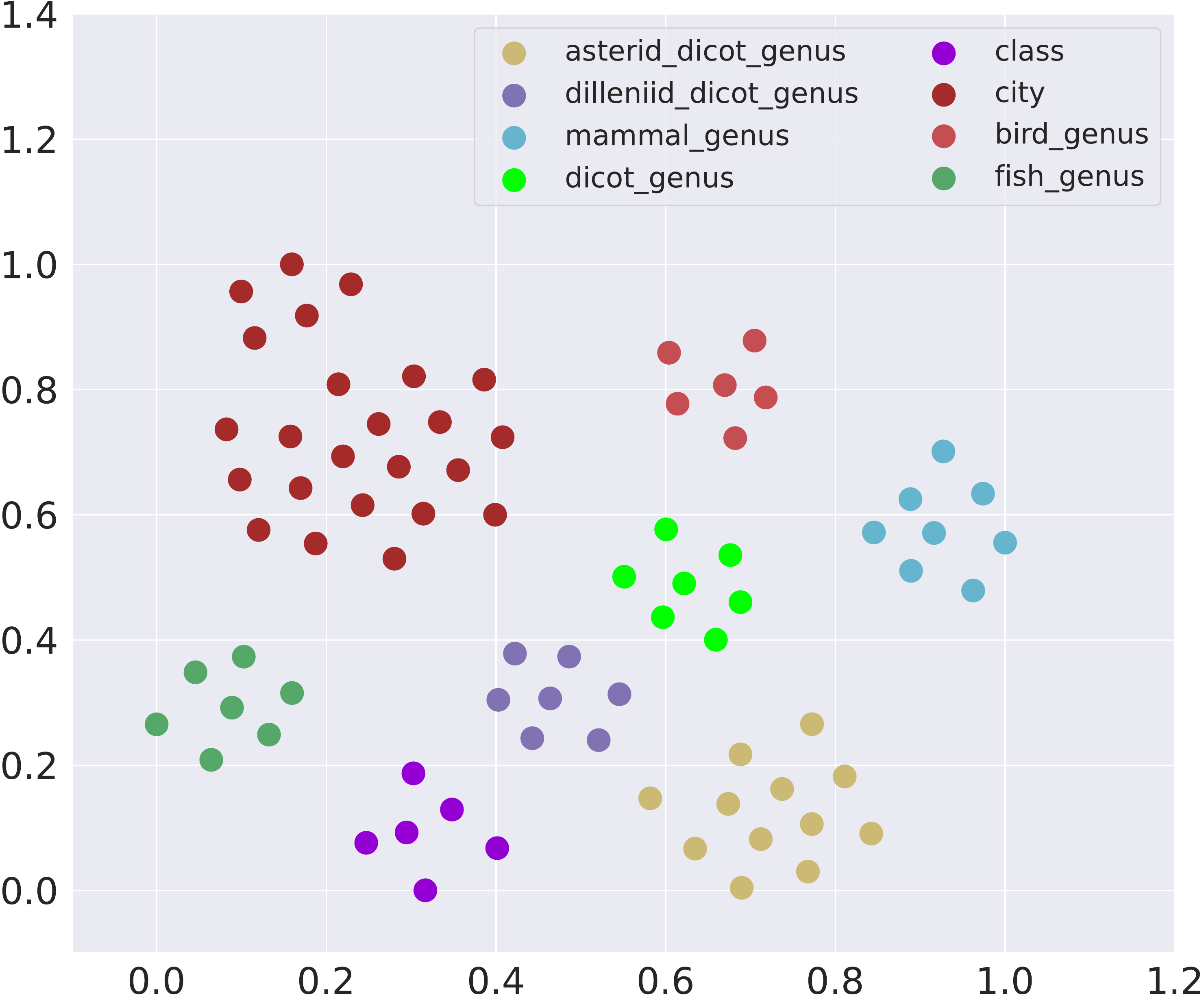}}
	\hspace{.1in} 
	\subfigure[RESCAL-CL]{\includegraphics[width=0.31\textwidth]{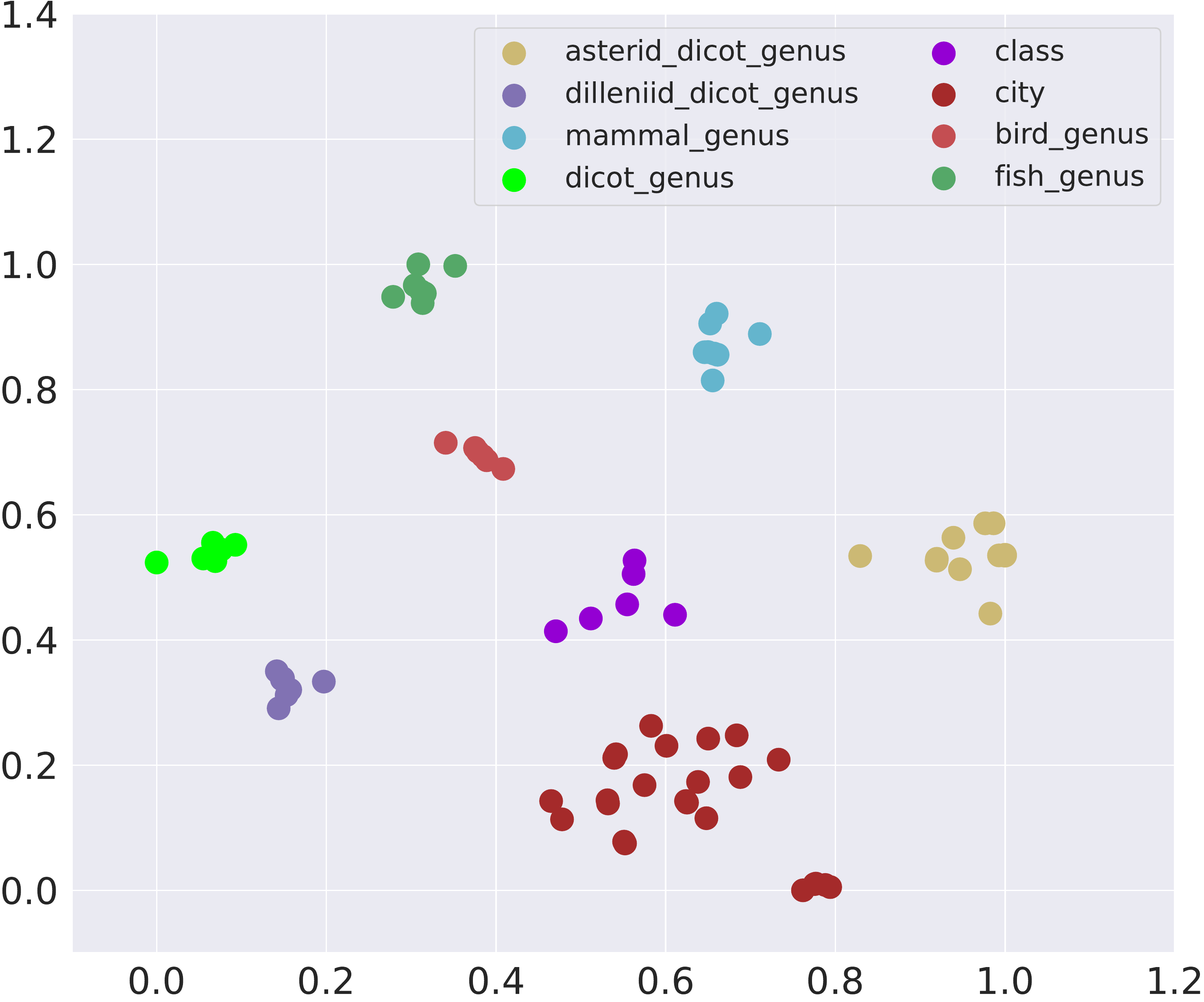}}
	\caption{The visualization of the entity-relation couples' embeddings using T-SNE. A points represents a $(h_i, r_j)$ couple, and the points with the same color are the couples that connected with the same tail entity.}
	\label{fig: visualization}
\end{figure*} 
\paragraph{Effect of Positive Instances}
We apply an ablation study to verify the effect of positive instances. The variants of w/o Pos are the variants of RESCAL-CL and ComplEx-CL, which remove all original and additionally added positive instances in a mini-batch when calculating the contrastive loss.
Table~\ref{table:posivitve_instance} shows the results of ablation study on WN18RR and FB15k-237 datasets.
From Table~\ref{table:posivitve_instance} we know the positive instance of KGE can significantly improve the performance of KGE.
Therefore, the positive instances we constructed are effective for the KGE.

\begin{figure}[t]
	\centering
	\subfigure{\includegraphics[width=0.49\columnwidth]{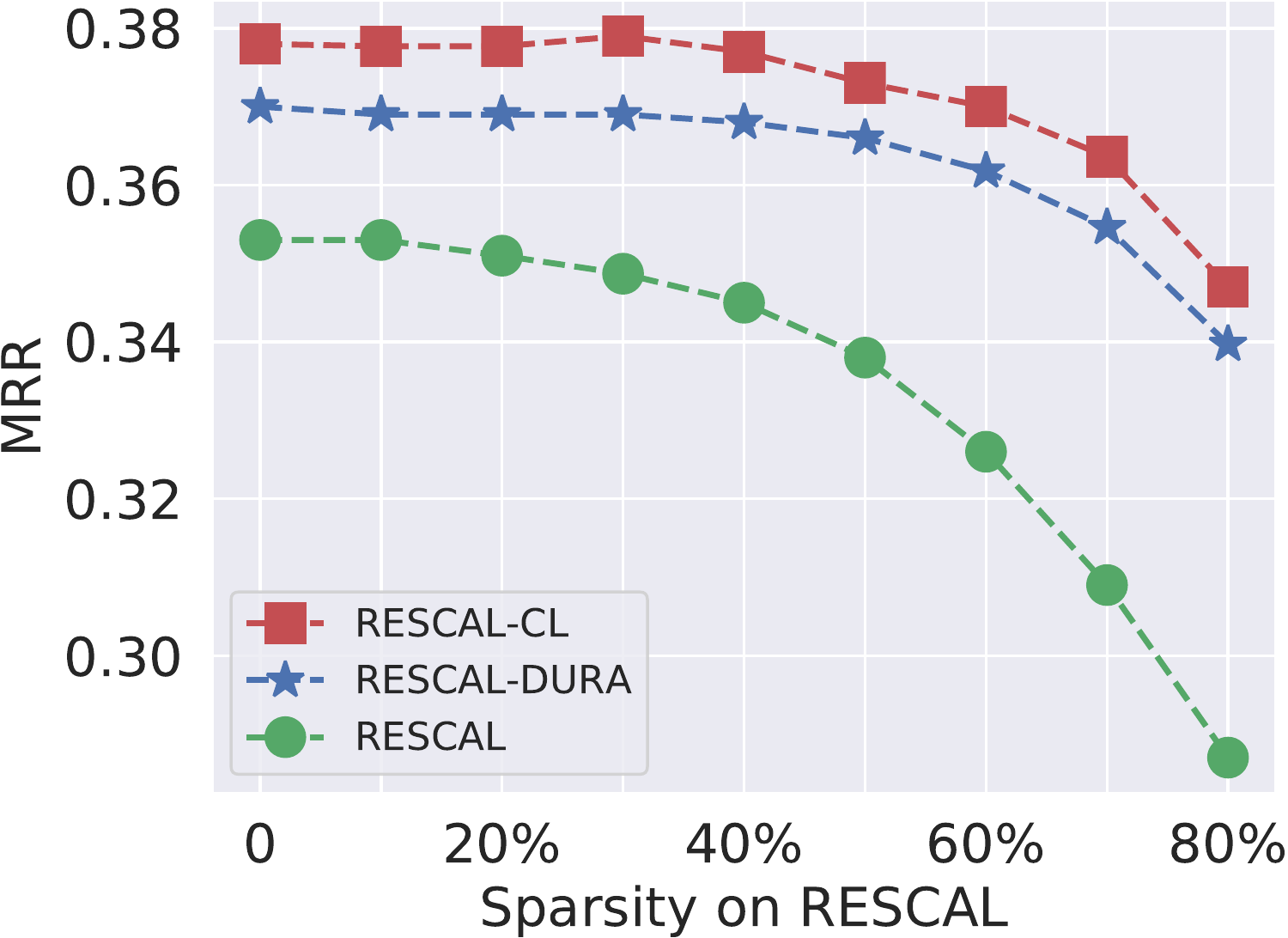}}
	\subfigure{\includegraphics[width=0.49\columnwidth]{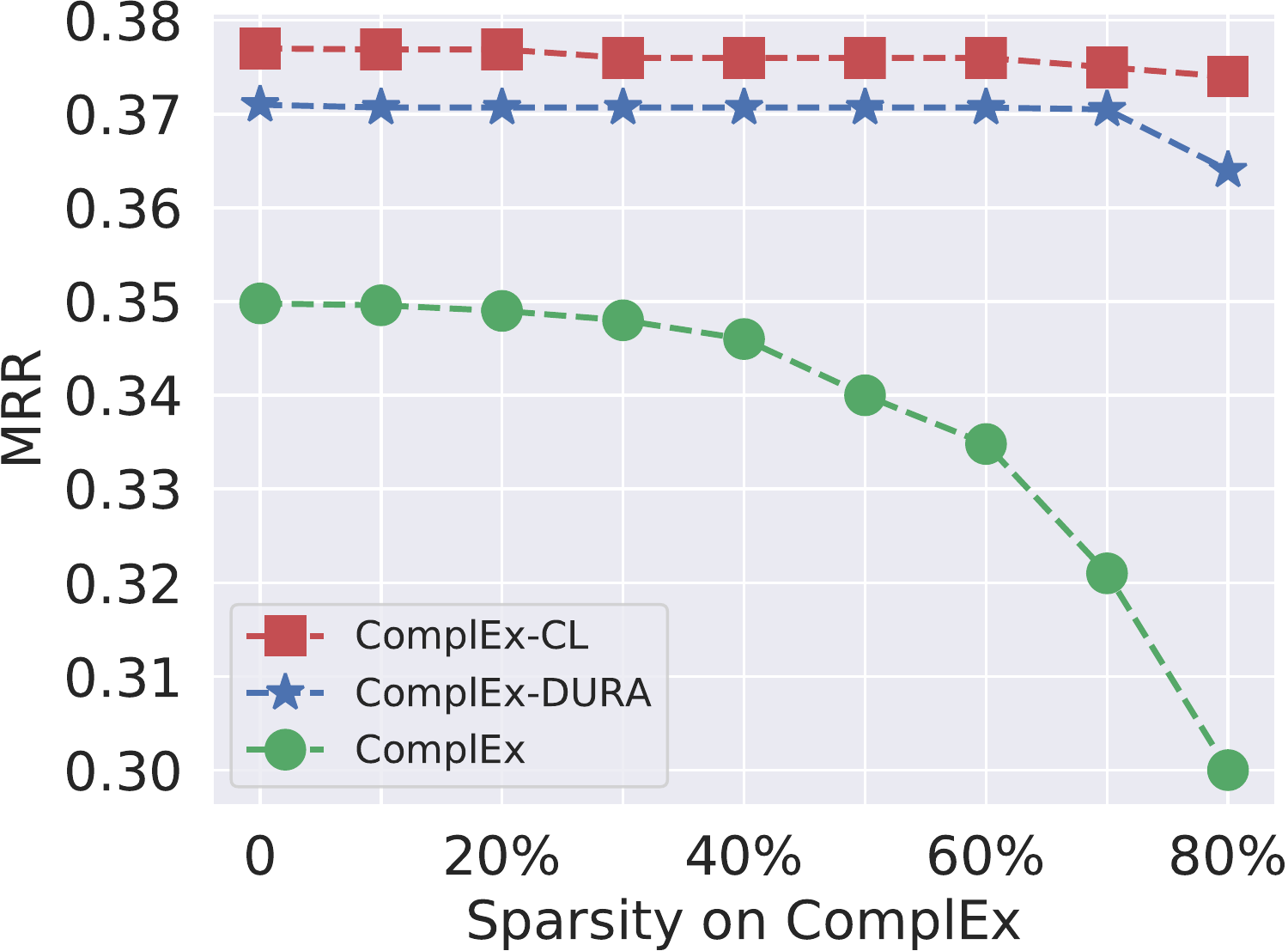}}
	\caption{Effect of sparsity on FB15k-237.}
	\label{fig:sparsity}
\end{figure} 
\paragraph{Sparsity Analysis}
As mentioned by some other work~\cite{dura}, the sparsity (the number of zero entries) of embeddings can save the storage usage of KG, which is vital for large scale real-world KGs.
Therefore, we analyze the sparsity of embeddings trainng by our contrastive learning (CL) framework.
We follow the setting in DURA~\cite{dura}, using a threshold $\lambda$ to mask a proportion of entries as zero, and observe the MRR results under different proportions.
Figure~\ref{fig:sparsity} shows the effect of embeddings' sparsity on FB15k-237. 
We can find that the embeddings trained by CL have better results than DURA in the sparse version, so CL can better reduce the storage usage of KG and benifit the real-world KGs.

\subsection{Visualization}
To make our method more explainable, we visualize the entity-relation couples via T-SNE~\cite{tsne}.
Specifically, we randomly pick up eight tail entities in WN18RR. 
We find out the triples with these tail entities in the test set, and extract $(h_i, r_j)$ couples in these triples. 
We visualize these couples' embeddings trained by the RESCAL, RESCAL-DURA, and RESCAL-CL.

Figure~\ref{fig: visualization} shows the results of visualization. The RESCAL method in Figure~\ref{fig: visualization} (a) can not properly separate the couples with different tail entities.
Compared with RESCAL, the RESCAL-DURA in Figure~\ref{fig: visualization} (b) can relatively better separate the couples with different tail entities.
However, since RESCAL-DURA can not capture the semantic similarity of couples with the same entity, the distribution of the couples connected with the same tail entity is still wide. 
Our RESCAL-CL can well split the couples in different types and shorten the distance of the couples connected with the same entity. Hence, our RESCAL-CL can better preserve the semantic information of the triples in knowledge graphs and has a higher performance.
	\section{Conclusion and Future Work}
\label{sec:conclusion}
In this paper, we propose a simple yet efficient contrastive learning framework for TDB KGE to improve its performance.
Compared with the previous work, our method can pull the related entities and entity-relation couples in different triples together in the semantic space and push the unrelated entities and couples apart.
The experimental results on the standard datasets show that our method can achieve new state-of-the-art results. Our analyses further verify the effectiveness of our approach.

In the future, we plan to extend the critical insights of contrastive learning to distance based (DB) KGE methods and other representation learning problems in natural language processing.
\section*{Acknowledgments}
Zhiping Luo, Wentao Xu and Jian Yin are supported by the National Natural Science Foundation of China (U1811264, U1811262, U1811261, U1911203, U2001211), Guangdong Basic and Applied Basic Research Foundation (2019B1515130001), Key-Area Research and Development Program of Guangdong Province (2018B010107005, 2020B0101100001).
\bibliographystyle{acl}
\bibliography{aaai22}
\end{document}